\documentclass{INTERSPEECH2023}

\usepackage{multirow}
\usepackage{color}
\usepackage[dvipsnames]{xcolor}
\usepackage[normalem]{ulem}
\usepackage{booktabs}
\usepackage[super]{nth}
\usepackage{makecell}
\usepackage{siunitx}
\usepackage{svg}
\usepackage{float}
\usepackage{amssymb}
\usepackage{url}
\usepackage{amsfonts}
\usepackage{pifont}%
\newcommand{\cmark}{\ding{51}}%
\newcommand{\xmark}{\ding{55}}%
\usepackage[caption=false]{subfig}
\usepackage{todonotes}
\usepackage{listings}
\newcommand{\bleu}{\textsc{BLEU}}
\newcommand{\covost}{\textsc{CoVoST}}

\interspeechcameraready

\title{Jointly Optimizing Translations and Speech Timing \\to Improve Isochrony in Automatic Dubbing}

\name{Alexandra Chronopoulou$^{1,2\hspace{0.5mm}*}$\thanks{*This work was done during an internship at Amazon.}, Brian Thompson$^{1}$, Prashant Mathur$^{1}$, Yogesh Virkar$^{1}$, \\  Surafel M. Lakew$^{1}$,  Marcello Federico$^{1}$ }
\address{$^1$AWS AI Labs\\$^2$Center for Information and Language Processing, LMU Munich, Germany}
\email{
\small \tt \{brianjt,pramathu\}@amazon.com}
\begin{document}
\ninept
\maketitle
\begin{abstract}
Automatic dubbing (AD) is the task of translating the original speech in a video into target language speech.
The new target language speech should satisfy isochrony; that is, the
new speech should be time aligned with the original video, including mouth movements, pauses, hand gestures, etc.
In this paper, we propose training a model that directly optimizes both the translation as well as the speech duration of the generated translations. We show that this system generates speech that better matches the timing of the original speech, compared to prior work, while simplifying the system architecture.

\end{abstract}


\section{Introduction} \label{sec:intro}

Automatic Dubbing (AD) \cite{oktem2019, federico_speech--speech_2020} is a specialization of speech-to-speech translation in which speech from a video is translated to a new language and the new speech is overlayed on the original video. 
One characteristic that sets dubs apart from other forms of translation is that the dubs should have a high \emph{isochrony}; 
that is, to the extent possible,
when a speaker's mouth is visible,
there should be speech in the dub language if the speaker's mouth is moving \cite{chaume:2004}.

In a perfect world, automatic dubbing systems would be trained on massive amounts of human-produced dubs,
and learn all aspects of dubbing (including isochrony) directly from the training data.
However, in practice, very little dubbing data is in the public domain due to the expense of creating dubbed content and it's commercial nature. 

Instead, research has focused on ways to produce isometric translations \cite{lakew2021isometricmt, isochronynmt} without relying on dubbing data for training,
and has largely focused on two areas: First, the generation of text translations such that, when synthesized into speech, the speech will be approximately the same duration as the source speech. This line of work has almost exclusively made the isometric assumption: that translations with approximately the same number of characters in as the source text will produce target speech that is the same duration as the source. The second line of work has been on modifying generated speech, via stretching/compressing phoneme durations \cite{virkar2021pa} during synthesis with text-to-speech (TTS) and adding pauses \cite{oktem2019pa,federico2020eval,virkar2021pa}, to make it's timing more closely match that of the source speech, often referred to as prosodic alignment (PA). These two approaches are typically used together to achieve high isochrony in target speech.

We identify two shortcomings of this standard approach:
\begin{enumerate}
    \item There is a little support for the isometric assumption used in prior isochrony aware MT work~\cite{isochronynmt,microsoftpaper}.
    In this work, we show that isometric MT is indistinguishable from standard MT in terms of isochrony.
    \item In cases where the MT system produces a translation which is too short or too long, PA is very limited in what it can do: it is separate from the MT model so it cannot change the translation to make it fit better, and it can't change the durations too much or the output speech will sound unnatural. 
\end{enumerate}

\begin{figure}[t]  \centering  \includegraphics[width=0.98\columnwidth, page=1]{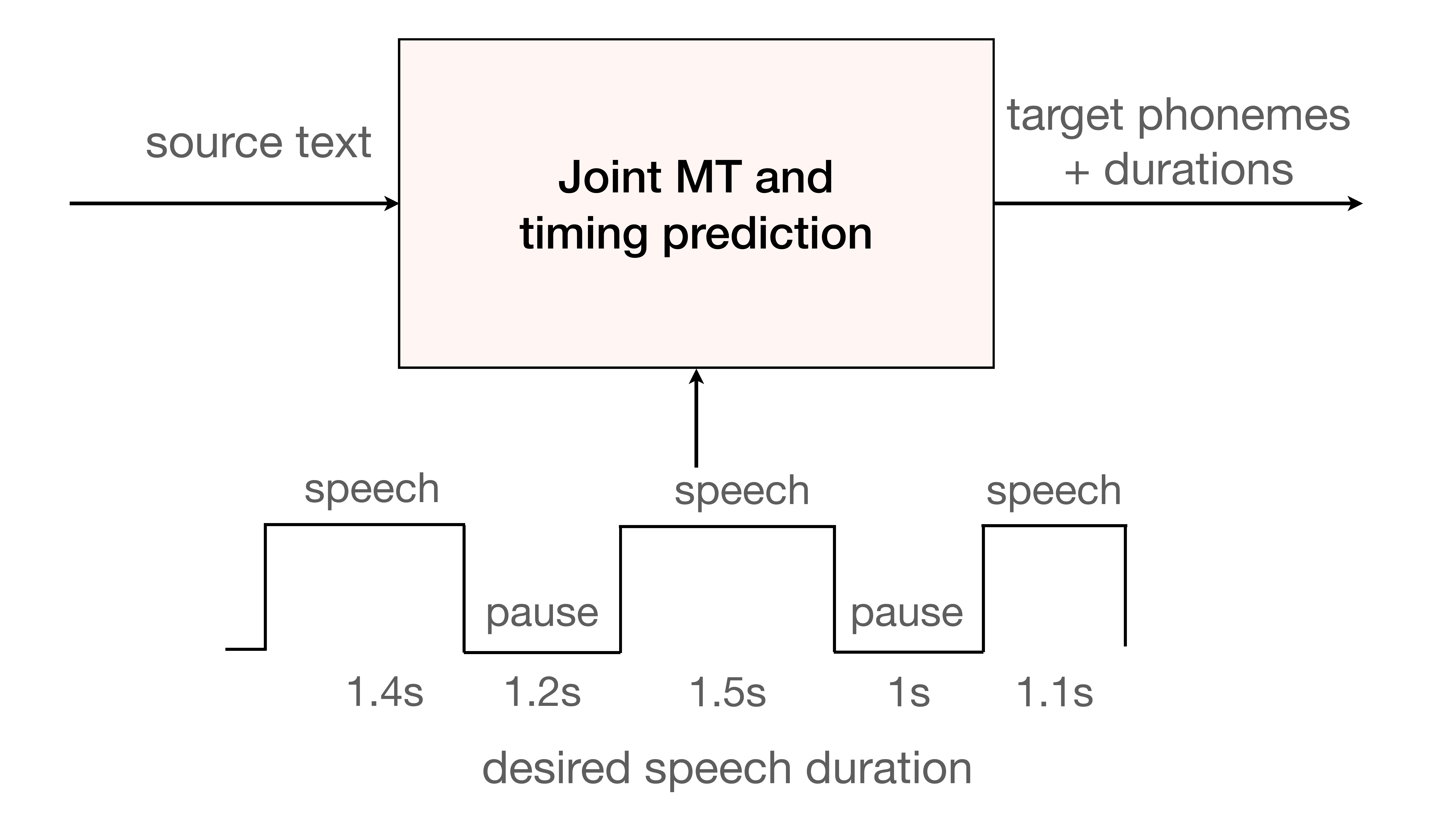}
    \caption{Our proposed model jointly translates and predicts the durations using a single encoder-decoder model.}
    \label{fig:model}
\end{figure}

To address these issues, we propose to \emph{jointly} optimize the translation and the timing of the translated speech. We achieve this by directly generating a representation used by many TTS models: phonemes and durations (see \autoref{fig:model}). 
Since we model durations explicitly, 
our method requires no assumptions about the duration that text will be after synthesis.
And since we jointly model translation and translation durations, the model is free to make sacrifices in translation quality to meet isochrony constraints and vice versa. 
In fact, we empirically show that our jointly optimized models achieve a significantly high speech overlap (i.e. isochrony) with a relatively low trade-off on translation quality.

To facilitate future research on automatic dubbing, in this paper we open source a new dubbing test sets in German-English translation direction with a CC-BY-4.0 license.

\section{Related Work}

 Most recent approaches on automatic dubbing ~\cite{oktem2019,federico_speech--speech_2020,federico_evaluating_2020, isochronynmt} are based on the following pipeline: a source speech utterance is passed through an ASR system, which provides a transcript in the source language. The transcript (in a pure textual form) is translated to the target language using an MT model, then a prosodic alignment model (PA) \cite{oktem2019, virkar2021pa} segments the translated text into phrases and pauses that follow the phrase-pause arrangement of the original speech  ~\cite{federico_evaluating_2020, virkar2021pa}. Finally, the output is passed to a PA module and finally to a TTS synthesizer. The TTS model adjusts the speaking rate of each phrase to make the speech segment approximately fit the timing of the corresponding source phrase \cite{virkar2021pa}.

To achieve isochrony, previous works have suggested training an MT system to generate translations whose length matches the length of the source transcript \cite{federico_speech--speech_2020, lakew2021verbosity, lakew2021isometricmt, isochronynmt}. Other works have applied re-scoring of $n$-best MT outputs either based on syllables~\cite{saboo-baumann-2019-integration} or length compliance~\cite{lakew2021isometricmt}. The intuition behind these works is isometry - a requirement that the number of characters/syllables in the target sentence is approximately the same as in the source sentence - which is a proxy of isochrony (similar duration). However, this assumption might not hold, as characters/words/syllables might have a different speech duration in different languages. Recent work did not find a strong correlation between isometry and isochrony in a human dubbed data \cite{brannon2022dubbing}, however it is unclear whether isometric assumption is a good proxy for automatic dubbing as \cite{microsoftpaper} observe improvements in isochrony scores for isometric system over the standard translation model in their human evaluations.
 
Concurrently to our work, \cite{microsoftpaper} propose designing a machine translation model specifically for automatic dubbing, by directly considering the speech duration of each word in translation to match the length of source and target speech. Our work differ from theirs in that we predict duration of the corresponding phonemes instead of words.

On the TTS front, there have been a few works addressing the dubbing problem. Neural Dubber~\cite{NEURIPS2021_8a9c8ac0} leverages a multi-modal setup to constrain the voice generation on the input text and the video. Prosodic alignment (PA) with relaxation has been used to fix the naturalness issue in the generated speech~\cite{virkar2021pa}, but both PA and TTS components used different duration models. \cite{effendi} proposed a single but shared duration model between PA and Tacotron~\cite{tacotron} (TTS model) to address the mismatch of separate duration models. In this work, we build on a similar hypothesis to~\cite{effendi}, in that we hypothesize that translation model should be aware of the durations of the target sequence it generates.

\section{Method}

We propose to jointly model translation and timing of translated speech. 
One of the main challenges in doing so is lack of obvious training data.
Human generated dubbing data is scarce and very rarely in the public domain 
due to the coast and commercial nature of dubbed media. 

\subsection{Training Data}\label{sec:training_data}

To address the lack of dubbing data, we propose to derive the training features we need from much more widely available \emph{speech translation} data. 
In particular, we select data where we have following triplet: 1) \textit{source text} 2) \textit{target speech} 3) \textit{target transcript}.

We compute the alignment between speech and text using Montreal Forced Aligner\footnote{\url{https://github.com/MontrealCorpusTools/Montreal-Forced-Aligner}} (MFA) \cite{mcauliffe17_interspeech}. MFA  provides phonemes and their durations on the target side and timestamps that allow us to detect pauses.\footnote{We define a pause to be silences of 300 ms or more.} This gives us the following:
1) \textit{source text} 2) \textit{target phonemes} 3) \textit{target durations} 4) \textit{locations of pauses in the target}. 

Given the target phoneme durations and the pauses, we compute speech segment durations  by adding up the phoneme durations between each pause. We train the MT model to take \textbf{source text} and \textbf{desired speech durations} as input, and predict \textbf{phonemes} together with their \textbf{durations} in the target.

\subsection{Binning of Speech Durations}

Desired speech durations (between each pause) are passed into the model using tokens appended to the source sentence.
Instead of using raw speech durations in seconds/frames, we instead bin them (group them together in groups that have equal number of samples). This is done so that each duration tag will be seen a reasonable number of times in training.
We group the durations into 100 bins. 

\subsection{Model Architecture}

We train a standard encoder-decoder Transformer model with source sequence as source text and the desired speech durations as bins (separated by a delimiter) and target sequence as phonemes and their durations in an interleaved way as shown below. We keep  end of word markers in the target phoneme sequence to retrieve the words back in target (natural) language and compute translation quality.

\begin{lstlisting}
Source: Lass es gut sein DELIM BIN4
Target: L 10 EH1 6 T 15 EOW W 4 EH1 6 L 7 EOW AH0 7 L 6 OW1 8 N 7 EOW
Gloss: Let well alone
\end{lstlisting}

During training, the desired timing information is derived from the target speech, in order to be able to train on data without source speech. 
We use cross-entropy loss in training in which phonemes and durations are weighted equally, i.e. we look at exact duration match in training time.\footnote{Durations do not need to be exact, but we leave this line of work as a future research direction.}
In inference, the timing is determined by the source speech and the model will be requested to match that timing in the target.

\subsection{Speech Synthesis}

Once we have target phonemes and phoneme durations, we use a publicly available FastSpeech 2 \cite{ren2021fastspeech} TTS model in order to generate speech. We provide the phonemes and phoneme durations as input to the model, overriding the native duration prediction model of FastSpeech 2.

\subsection{Noise to Address Training/Inference Mismatch}
\label{subsec:noise}

In training, we use force alignment on the target speech and transcript to determine the phonemes and their durations, and this results in near perfect durations on target side. 
In inference, we run a Voice Activity Detector (VAD)~\cite{SileroVAD} on the source speech to get the speech segments (around the pauses) and their durations.
These durations from VAD in act as desired target durations  which is a much weaker signal for the model given that it has seen near perfect duration in training.
To address this mismatch we add noise in the source durations, specifically we add Gaussian noise (based on standard deviation) as our initial analysis suggest a Gaussian distribution of durations in training.

\section{New Dubbing Test Set} 
\label{sec:dubbing}
Most of the public dubbing data sets like Heroes~\cite{oktem2018heroes}, real world dubbing data~\cite{microsoftpaper} cannot directly be used for research because they either contain copyrighted material or are not reproducible.
In order to evaluate our proposed pipeline and facilitate further dubbing research, we created a true dubbing data (with videos) which can be used by the research community under Apache 2.0 license.\footnote{Dubbing test set released here \url{https://github.com/amazon-science/iwslt-autodub-task}}
We created two dubbing test sets which are extracted from the En$\rightarrow$De test set of \covost-2 \cite{covost1}, a large-scale multilingual speech translation corpus based on Common Voice.\footnote{\url{https://commonvoice.mozilla.org/}} 

Specifically, the first subset is created by randomly sampling 91 sentences (\textit{test91}), while for the second, we randomly sampled 101 sentences from the longest 10\% of the De part of the test set (\textit{test101}). We made sure that there is no overlap between the two subsets. We annotated the German side of \textit{test101} with pause tokens, added in places where a native speaker would pause (for more than $0.3$s). Each sentence in \textit{test101} has 1 or more pause tokens. 

Then, we asked volunteers (German speakers) to read the sentences from the two subsets, creating one clip for each sentence. For \textit{test91}, they were instructed to speak naturally and pause if/when they feel it is natural; for \textit{test101}, they were instructed to speak naturally and pause when they see a pause token. The face of each speaker is visible in each clip, there is minimal background noise and they are directly facing the camera. We perform human evaluation using these two subsets.

\begin{table}[h]
\centering 
\begin{tabular}{lrrr}
\toprule
 & \textbf{Train} & \textbf{Valid} &  \textbf{Test}  \\  \midrule 
\# samples  & 289,074 & 15,499 & 15,413 \\ \midrule 
\# samples with 1+ pauses & 36,183  & 2,662  & 2,632  \\
\% samples with 1+ pauses & 12.5\% & 17\%  &17\% \\ \midrule 
\# samples with 2+ pauses & 5,490  & 486  & 569 \\
\% samples with 2+ pauses & 2\% & 3\%  & 4\% \\
\bottomrule   
\end{tabular}
\caption{Dataset sizes and statistics of the pauses in the dataset. A pause is defined as at least 300ms of silence between two consecutive spoken words, following \cite{virkar2021pa}.}
\label{table:statistics}
\end{table}

\section{Experiments}

\subsection{Datasets and Processing}

\noindent \textbf{Training Data.} 
We use the English-German portion of the \covost-2 training set, i.e. all data for which English speech (along with transcripts) and German text translations are available. 
However, as described in Sec. \ref{sec:training_data} we require target speech and transcript to train our models, as such we flip the language direction to German-English where we have English speech and transcript in place. 

In Table \ref{table:statistics}, we present some details about the dataset we used (\covost-2). Specifically, we show the number and percentage of sentences which have pauses. We observe that only a small fraction of the training dataset (12.5\%) has any pauses at all. This is probably the case because most samples (sequences) in the dataset consist of only 1 sentence, and are therefore quite short. Specifically 98.6\% of the samples in the training dataset (and 98.5\%, 98.3\% of the validation and test datasets respectively) consist of a single sentence. We evaluate our approach and the baselines on the \covost-2 test set as well as the two dubbing test sets \textit{test91} and \textit{test101}.

\noindent \textbf{Preprocessing.} 
We use Byte-Pair-Encoding (BPE)~\cite{sennrich-etal-2016-neural} to split the source and target side of the training data with a vocabulary of 10K tokens when both source and target contain text. 
If there are phonemes in the target side, we use a closed vocab. %
We generate speech using an open-source implementation of FastSpeech 2~\cite{ren2021fastspeech}. We use the default hyper-parameters for pitch and energy prediction, but override the duration predictor with the ones predicted by our models. 

\subsection{Model and Training Configuration}
We use the base configuration of the Transformer \cite{vaswani2017attention} with $6$ encoder and decoder layers, self-attention dimension $512$ and $8$ heads, and feed-forward sublayers of dimension $2048$. The model is optimized with Adam \cite{kingma2014adam}, with an initial learning rate of $5 \times 10^{-4}$. Dropout is set to $0.3$. The best model from each configuration is selected based on the checkpoint that has the lowest validation BLEU. We train each model for $200$ epochs and use the best checkpoint (based on the dev set) for evaluation. At inference time, we use a beam size of  $5$ for decoding and evaluate BLEU scores~\cite{papineni2002bleu} using Sacre$\bleu$ \cite{post-2018-call}.\footnote{``nrefs:1$|$case:lc$|$eff:no$|$tok:none$|$smooth:exp$|$version:2.2.1''} We use the fairseq library for all our experiments~\cite{ott-etal-2019-fairseq}. To generate speech, we use FastSpeech 2 for which a publicly available pretrained TTS English model is available.\footnote{Fork with changes to the original package: \url{https://github.com/mtresearcher/FastSpeech2}}

\subsection{Baselines \& Models}

We compare our approach against three baselines. The first is a standard NMT model, named \textit {StdMT}, trained to translate text into text. The second baseline, named \textit{IsoMT} , is the isometric NMT model by \cite{lakew2021isometricmt} that is trained to generate translations that match the input length in terms of number of characters.  The third baseline, named \textit{Txt2Phn}, is closely related to our approach as it trains a model with source text (without durations) as input, and target phonemes and phoneme durations as output. All baselines are trained using the same hyperparameter configuration.

We train our proposed approach, \textit{Txtd2PhnD},  with source text and durations (obtained from the target speech) as input, and target phonemes and phoneme durations as output. As explained in Section~\ref{subsec:noise}, we also experiment with our model by adding increasing amounts of Gaussian noise (with standard deviation ranging from 0.1 to 1.5) to the source durations.

\begin{figure}[h]
    \centering
    \includegraphics[width=1\columnwidth, page=1]{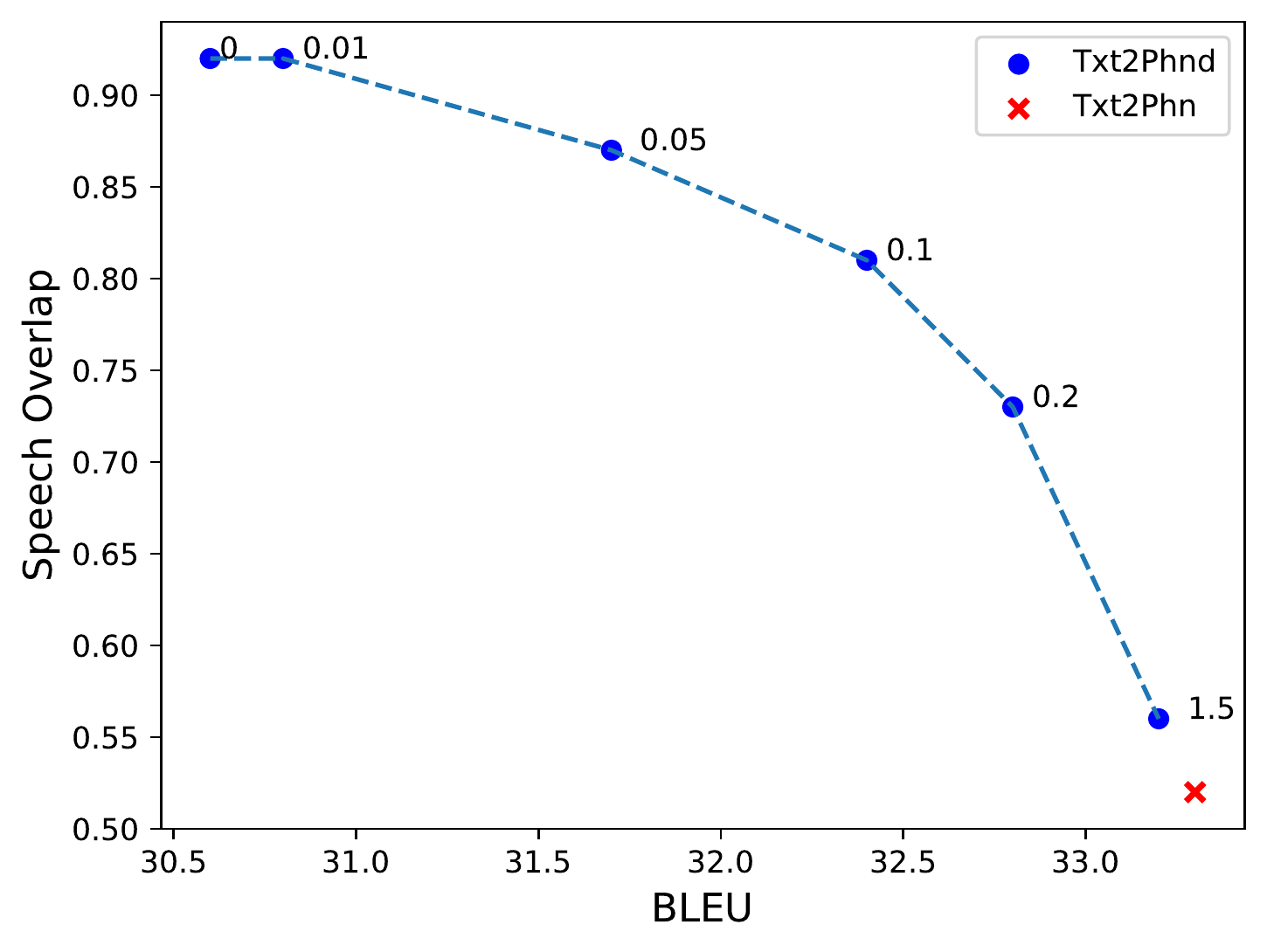}
    \caption{Test BLEU and speech overlap. We increase the noise added to the source speech durations (standard deviation shown on the data points). We notice a trade-off between the synchronicity of the system and its translation quality. }
    \label{fig:overlapbleu}
\end{figure}

\subsection{Evaluation Metrics}

One challenge in evaluation is the fact that we are generating phonemes instead of words. We initially tried measuring phoneme overlap but found this 
penalized the difference between valid alternative phonetic transcriptions of the same word (e.g. \textit{AE1 N D} and \textit{AH0 N D} for "and").

To avoid such penalties, we instead chose to map phoneme sequences back to words. As there is no one-to-one mapping from phonetic transcriptions to words, e.g. \textit{K AE1 T S} is the transcription of both ``cat's'' and ``cats'', we trained a phonemes-to-words seq2seq ``MT'' system, using the English \covost-2 training data. We formed our parallel data using force-aligned phonemes in the source and text in the target. We used this trained model to evaluate our proposed approach for translation quality. 

To measure the synchronicity between source and dubbed speech, we consider the following metric:
\begin{align}
    \textit{speech overlap} &= 1 - \frac{ | \textit{source duration} - \textit{dub duration} |}{\textit{source duration}}   
\end{align}

We take the difference between the durations of source and dubbed speech and normalize it over the source durations to not have any bias over the length of the segment.
We then compute the overlap at a segment level and report the average across the test set.

\section{Results}

\subsection{Automatic Evaluation Results}

 \begin{table}[t]
\centering 
\begin{tabular}{lcccccc}
\toprule
\multirow{2}{*}{\textbf{System}} & \multirow{2}{*}{\textbf{Ph.}} & \multicolumn{2}{c}{\textbf{Durations}} &  \multirow{2}{*}{\textbf{BLEU} $\uparrow$} & \multirow{2}{*}{\textbf{SO} $\uparrow$}  \\
 & & \textbf{I/P}  & \textbf{O/P} &  &  \\  \hline
{\textit StdMT} & \xmark &  \xmark       & \xmark & 35.6 & 0.53   \\  
\textit{IsoMT} &  \xmark &   \xmark       & \xmark & 35.3 & 0.53  \\  \hline
\textit{Txt2Phn} &  \cmark & \xmark       & \xmark & 33.3 & 0.52  \\ 
\textit{Txtd2PhnD}  &   &  &  &  \\ 
\hspace{6pt} +w/o noise  & \cmark & \cmark      & \cmark & 30.6 & 0.92 \\
\hspace{6pt} +w/ 0.01 noise  & \cmark & \cmark      & \cmark & 30.8 & 0.92 \\
\hspace{6pt} +w/ 0.1 noise  & \cmark &  \cmark      & \cmark & 32.4 & 0.81 \\
\bottomrule 
\end{tabular}
\caption{Results for De $\rightarrow$ En on \covost-2 test set. We evaluate our system both in terms of translation quality (BLEU) and speech overlap (SO). \cmark~ and ~\xmark~ denotes whether target text contain phonemes or text (\textbf{Ph.}) and/or if there are durations provided in source (\textbf{I/P}) or predicted in target side (\textbf{O/P}).}
\label{table:resultscovost}
\end{table}

\noindent \textbf{Full Test Set:}  Results of all systems evaluated on full \covost-2 test set (~15K samples) for De-En are shown in Table \ref{table:resultscovost}. We present results both in terms of translation quality and speech overlap. 

In terms of speech overlap, we see that the speech produced baseline models (\textit{StdMT}, \textit{IsoMT}, and \textit{Txt2Phn}) have low speech overlap with the source speech (with values $\approx$0.5) because the target speech durations are independently predicted by FastSpeech2 which does not have information of source side speech durations. In contrast, our proposed set of models \textit{Txtd2PhnD} - where they have access to phrase level source speech durations - produces significantly more synchronized target speech (55\% relative) with minimal trade off on translation quality (3\% to 9\% relative).
This intuitively makes sense, as we expect that the best dubbing speech is not necessarily obtained from the best stand-alone translation; instead, the generated output needs to both be similar in terms of content with the source but also match the pauses and prosodic structure of the source speech.

Directly optimizing durations and translation creates an inherent trade-off which is depicted in Figure \ref{fig:overlapbleu}. Specifically, if we add Gaussian noise of increasing standard deviation to the source speech durations (and oversample the dataset, so that a single text sequence receives different speech durations in different batches), we observe that the translation quality monotonically increases, while the speech overlap decreases. Depending on the importance of speech overlap for a specific application, we can adjust the amount of noise and obtain a model that translates reasonably and generates speech that is temporally well aligned with the source. 
 
\noindent \textbf{New Dubbing Test Set:} In Table \ref{table:resultsnewdubbing}, we compare same set of systems as above on the two dubbing test sets. The trends we observe are similar: our model provides translations with higher overlap but with lower BLEU scores. Compared to \textit{Txt2Phn}, which is arguably the closest baseline, our model generates speech that has a higher isochrony. The performance improvement is more evident in \textit{test91} (up to $+ 0.29$ in terms of speech overlap) than \textit{test101} (up to $+ 0.17$) which is a harder test set of the two as it contain pauses.

We observe that isometric MT (in isolation, without prosodic alignment) is no more isometric than standard MT when both are fed into the same TTS model, calling into question whether it is helpful in dubbing at all. However, it is possible that isometric MT interacts more positively with prosodic alignment. 

\begin{table}[t]
\centering 
\begin{tabular}{lcccc}
\toprule
\multirow{2}{*}{\textbf{System}} & \multicolumn{2}{c}{\textit{test91}} & \multicolumn{2}{c}{\textit{test101}}  \\
 & \textbf{BLEU} $\uparrow$ & \textbf{SO} $\uparrow$ & {\textbf{BLEU}} $\uparrow$ & {\textbf{SO}} $\uparrow$ \\  
\midrule 
{\textit Std MT} & 38.6   & 0.54 & 28.7 & 0.32  \\  
{\textit IsoMT} & 40.3  & 0.54 & 29.5 & 0.33  \\ 
\textit{Txt2Phn} & 34.4 & 0.56   & 28.0 & 0.53  \\ \hline
\textit{Txtd2PhnD} & & & &  \\ 
\hspace{6pt} w/o noise & 27.2 & 0.84 & 23.8 & 0.7 \\ 
\hspace{6pt} w/ noise 0.01 & 27.3 & 0.81 & 25.0 & 0.68 \\ 
\hspace{6pt} w/ noise 0.1  & 31.7  & 0.59 & 27.2 & 0.62  \\ 
\bottomrule   
\end{tabular}
\caption{Results with translation quality (BLEU) and Speech Overlap (SO) for De $\rightarrow$ En language pair on two dubbing test sets \textit{test91} and \textit{test101}.}
\label{table:resultsnewdubbing}
\end{table}

\section{Conclusion}
We introduce a novel method that jointly optimizes a MT model for translation quality and speech duration overlap with the targeted use case of automatic dubbing. 
We present a model that generates translations at phoneme level, along with their durations, thus resulting in a much simpler AD pipeline which produces natural dubbed videos. 
Empirical studies of our proposed models show significant gains of 55\% relative in speech overlap over both \textit{Txt2Phn} and \textit{StdMT} baselines with a trade-off of 2.7\% and 9\% in translation quality respectively.
We also release a dubbing test set which can be used by researchers in the dubbing community. 

\bibliographystyle{IEEEtran}
\bibliography{paper}

\end{document}